\documentclass[letterpaper, 10 pt, conference]{ieeeconf}  

\IEEEoverridecommandlockouts                              

\usepackage{graphics} 
\usepackage{epsfig} 
\usepackage{times} 
\usepackage{amsmath} 
\usepackage{amssymb}  
\usepackage{algorithm}
\usepackage{algorithmic}

\usepackage{cite}
\usepackage[]{hyperref}

\usepackage{xcolor}

\title{\LARGE \bf
Vibration-Based Energy Metric for Restoring Needle Alignment in Autonomous Robotic Ultrasound
}

\author{Zhongyu Chen$^{1,\dagger }$, Chenyang Li$^{2,\dagger}$, Xuesong Li$^{3,6}$, Dianye Huang$^{3,6}$, Zhongliang Jiang$^{3,6}$, \\ Stefanie Speidel$^{2,5}$, Xiangyu Chu$^{1,4,*}$, and Kwok Wai Samuel Au$^{1,4}$
\thanks{This work was supported in part by the Multiscale Medical Robotics Centre, AIR@InnoHK, in part by
Direct Grants (The Chinese University of Hong Kong) under Grant 4055245, in part by SINO-German Mobility Project under Grant M0221, in part by the Federal Ministry of Research, Technology and Space of Germany through grant 01IS23070 Software Campus 3.0 TUD Dresden University of Technology as part of the Software Campus project 'NeuralNodes', and in part by the project “Next Generation AI Computing (GAIn),” from the Bavarian Ministry of Science and the Arts and the Saxon Ministry for Science, Culture, and Tourism.
$^{1}$Multi-scale Medical Research Center, Hong Kong, China. 
$^{2}$ Department of Translational Surgical Oncology, National Center for Tumor Diseases (NCT/UCC), Dresden, Germany; German Cancer Research Center (DKFZ), Heidelberg, Germany; Faculty of Medicine and University Hospital Carl Gustav Carus, TUD Dresden University of Technology, Dresden, Germany; Helmholtz-Zentrum Dresden-Rossendorf (HZDR), Dresden, Germany. 
$^{3}$ Chair for Computer Aided Medical Procedures and Augmented Reality (CAMP), Technical University of Munich, Munich, Germany.
$^{4}$Department of Mechanical and Automation Engineering, The Chinese University of Hong Kong, Hong Kong, China.
$^{5}$ Centre for Tactile Internet with Human-in-the-Loop (CeTI), TUD Dresden University of Technology, Dresden, Germany.
$^{6}$ Munich Center for Machine Learning (MCML), Munich, Germany.
 $\dagger$: Equal contribution; $*$: Corresponding author.
}
}

\begin{document}

\maketitle
\thispagestyle{empty}
\pagestyle{empty}

\begin{abstract}

Precise needle alignment is essential for percutaneous needle insertion in robotic ultrasound-guided procedures. However, inherent challenges such as speckle noise, needle-like artifacts, and low image resolution complicate robust needle detection, which is essential for alignment in ultrasound images. These issues become particularly problematic when visibility is reduced or lost, diminishing the effectiveness of visual-based needle alignment methods. In this paper, we propose a method to restore effectively when the ultrasound imaging plane and the needle insertion plane are misaligned. Unlike many existing approaches that rely heavily on needle visibility in ultrasound images, our method uses a more robust feature by periodically vibrating the needle using a mechanical system. Specifically, we propose a new vibration-based energy metric that remains effective even when the needle is fully out of plane. Using this metric, we develop an elegant control strategy to reposition the ultrasound probe in response to misalignments between the imaging plane and the needle insertion plane in both translation and rotation. Experiments conducted on ex-vivo porcine tissue samples using a dual-arm robotic ultrasound-guided needle insertion system demonstrate the effectiveness of the proposed approach. The experimental results show the translational error of 0.41$\pm$0.27 mm and the rotational error of 0.51$\pm$0.19 degrees.
\end{abstract}


\section{INTRODUCTION}
Precise percutaneous needle insertion is crucial for many minimally invasive procedures, including tissue biopsy and ablation, ensuring precision and procedural success~\cite{kuang2016modelling,masoumi2023big}. 
Due to its radiation-free nature and real-time imaging, medical ultrasound (US) is widely used to guide needle punctures. However, traditional free-hand ultrasound examination is highly operator-dependent, leading to intra- and inter-operator variations\cite{jiang2024intelligent}. Additionally, training qualified sonographers is time-intensive. To address this, many Robotic Ultrasound Systems (RUSS) have been developed to improve needle intervention accuracy~\cite{jiang2023robotic}. Although robotic assistance does not directly improve needle visualization, it enhances needle control and localization accuracy~\cite{beigi2021enhancement}.

\begin{figure}[t]
\centering
\includegraphics[width=0.45\textwidth]{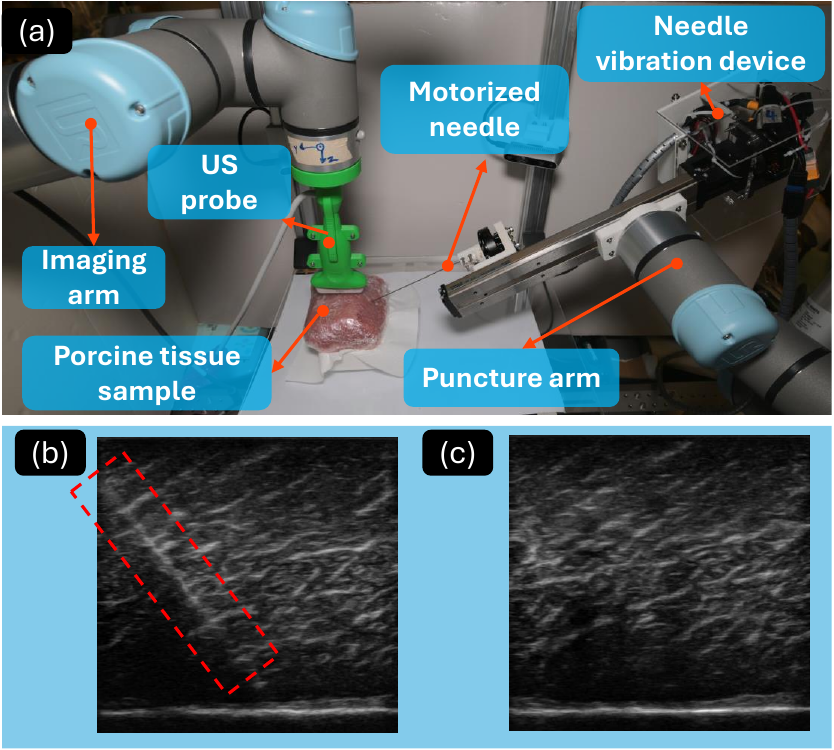}
\caption{Illustration of dual-arm robotic needle insertion on an ex-vivo porcine tissue sample. (a) demonstrates data acquisition setup on an ex-vivo animal tissue sample. (b) is a typical B-mode image with an in-plane clear needle, and (c) is the B-mode image with an invisible needle acquired at the position where the probe is parallel to the plane shown in (b) at a distance of 3.6 mm. 
}
\label{Fig_platform}
\end{figure}



Nevertheless, the practical use of RUSS for needle insertion still faces significant challenges. It not only requires accurate needle detection in the US images but also faces the difficulty of keeping the alignment between the imaging and the insertion plane. To solve these challenges, several works proposed visual servoing methods to restore the alignment using information from needle detection in the US images. Chatelain et al.~\cite{chatelain2013real} detected a needle in 3D US images to keep the needle tip centered in the image and the needle axis aligned with the longitudinal axis of the probe. Jiang et al.~\cite{jiang2024needle} proposed a framework to address the misalignment of needle and probe based on the needle segmentation in 2D US. The performance of their needle alignment methods is highly dependent on visual features, demanding high-precision needle detection. 

However, inherent limitations of US imaging, such as speckle noise and acoustic artifacts, make continuous needle tracking within the 2D US plane difficult during insertion~\cite{reusz2014needle}. Additionally, due to the beam profile constraints of 2D US, needle bending and minor movements can lead to misalignment with the imaging plane, further complicating accurate detection and precise puncture~\cite{bi2024machine, jiang2022towards}. For example, in Fig.~\ref{Fig_platform} (b) and (c), the needle disappears from the images with just a 3.6 mm translational offset, demonstrating that relying solely on visual features for needle guidance in RUSS is impractical. Even when properly aligned within the 2D US imaging plane, poor needle visibility remains a common issue. This is due to the mirror-like reflective nature of standard metal needles, where specular reflection reduces echogenicity and diminishes visibility~\cite{beigi2021enhancement,malamal2023enhanced}. Therefore, many visual-based needle alignment methods may lose their effectiveness when the needle is not visible, making robust needle detection an important preliminary task.



To enhance needle detection in 2D US images, various approaches have been explored, broadly categorized into hardware-based, software-based, and hybrid methods. Among hardware-based approaches, some use electromagnetic or optical trackers to track the needle~\cite{stolka2014needle} while others enhance needle visibility through specialized designs~\cite{lu2014new, hovgesen2022echogenic} or employ external needle guides to facilitate puncture~\cite{prasad2015real}. Although these methods are effective in certain scenarios, their high cost and clinicians' preference for standard needles make the adoption of these customized apparatuses challenging in clinical practice~\cite{beigi2021enhancement}. Another possible hardware-based solution is using a 3D US probe, which has been explored to enhance needle visualization by providing additional spatial information. However, its application remains limited due to constraints such as reduced image quality and higher computational complexity~\cite{jiang2024needle}.
Software-based methods primarily detect the needle using B-mode US images or radio frequency (RF) data for enhanced visualization and tracking. Various classical computer vision techniques have been applied in this domain, including the Hough transform~\cite{okazawa2006methods}, the Log-Gabor wavelets~\cite{hacihaliloglu2015projection}, and the Gabor filter plus Random Sample Consensus (RANSAC)~\cite{kaya2014needle}. Lee et al.~\cite{lee2020ultrasound} incorporated ``Squeeze and Excitation" blocks to enhance feature representation, while Mwikirize et al.~\cite{mwikirize2021time} introduced a time-domain approach using convolutional neural networks (CNN) with Long Short-Term Memory (LSTM) modules to improve the needle detection. Chen et al.~\cite{chen2022automatic} proposed W-Net, a dual-branch network for automatic needle segmentation. Jiang et al.~\cite{jiang2024needle} leveraged GAN-based methods to enhance needle segmentation, and Wijata et al.~\cite{wijata2024needle} applied Vision Transformers (ViT)~\cite{dosovitskiy2020image} for more robust needle detection. However, in practice, these appearance-based methods fail to fundamentally resolve the needle invisibility issue caused by specular reflection and misalignment.
%

To overcome the aforementioned challenges, a series of hybrid methods combining hardware and software algorithms have been proposed. One such approach is the use of a vibrating needle. The vibrating needle could introduce additional features for enhancing needle visibility in US images. Previous studies have explored this approach: Harmat et al.~\cite{harmat2006needle} proposed a method utilizing controlled needle oscillation, demonstrating improved needle tracking accuracy. To magnify the tiny motion in US images, Huang~\emph{et al.} apply the video motion magnification technique on the relatively low-frequency US video recordings~\cite{huang2023motion}. Fronheiser et al.~\cite{fronheiser2008vibrating} investigated how needle vibration improves echogenicity, making it easier to detect in B-mode US images. Orlando et al.~\cite{orlando2023power} introduced combining B-mode and Doppler modes to detect vibrational motion, further enhancing needle visualization in US imaging. Although these methods have made significant progress in needle detection, they rely heavily on explicit handcrafted features and involve high computational costs, limiting their practicality for real-time applications. To address these limitations, Huang et al.~\cite{huang2025vibnet} proposed a novel end-to-end vibration-based needle detection method that simultaneously captures temporal and frequency features through specialized needle prediction modules. By using neural networks, the method accelerates inference to 12 Hz, enabling real-time performance and potential integration with RUSS for improved needle detection and guidance. 

Inspired by these vibration-based methods, we propose a new method that can restore the needle alignment using vibration. Our method is not highly dependent on the visibility of the needle in the US images but uses a more robust feature introduced by vibrating the needle periodically using a mechanical system. Thereby, it can maintain its effectiveness even when thin objects are fully out of the plane. Our main contributions include: 
\begin{itemize}
    \item We propose an effective vibration-based energy metric, which is a robust feature for restoring needle alignment. With this feature, the restoration can be done even when the needle is not visible in the US images. 

    \item For the US probe repositioning, an elegant robot control method is presented to restore the probe to the needle plane autonomously when the needle is not in the optimal imaging plane. The method is especially valuable for developing an autonomous robot for ultrasound-guided percutaneous needle insertion.

  \item The effectiveness of our proposed method is validated by experiments on porcine tissue samples with different insertion angles and depths. The experimental results show that our method can accurately restore the ultrasound probe to the optimal imaging plane.
\end{itemize}
The remainder of this paper is organized as follows. Section II details our proposed method for vibration-based energy analysis. Section III elaborates on our US probe reposition control method. Our current results are described in Section IV. Finally, we conclude our work and provide discussions in Section V.


\section{VIBRATION-BASED ENERGY ANALYSIS}


\subsection{Vibration-based Energy Heatmap}
Our dual-arm robotic needle insertion system is shown in Fig.~\ref{Fig_platform} (a). One robotic arm is connected to the US probe, while the other is combined with a needle that is vibrating by a motor. The needle's vibration can provide additional information, except for US imaging information. Based on this hardware system, we are allowed to study the mechanism of needle vibration and its application to needle alignment, mainly due to the controllable vibrating frequency and amplitude.

Previous work from Huang \emph{et al.}~\cite{huang2025vibnet} indicates that a significant vibrating frequency of the needle in the frequency spectrums can be observed from both the pixel inside the needle and the neighboring pixels, while no dominant frequency is detected from the pixels far from the needle. In addition, pixels adjacent to the needle exhibit a higher level of noise in the frequency spectrum compared to those within the needle.

Inspired by this observation, applying a bandpass filter to isolate intensity changes at frequencies near the vibration frequency and analyzing the energy of the remaining signals should reveal higher energy levels when the US probe is closer to the needle plane compared to when it is farther away.
To verify this hypothesis, we collect $T$ consequential US images $I\in(H, W)$, where $H$ and $W$ are the height and width, respectively. The bandpass filter is applied to each pixel's intensity change signal $\mathbf{S}$ as follows:
\begin{equation}
    \mathbf{S'}_{i,j} = iFFT(\mathbf{M} * FFT(\mathbf{S}_{i,j})),
\end{equation}
where the index of the pixel within an image is denoted by $i$ and $j$, $\mathbf{M}$ is the binary signal mask used to filter signals with frequencies in the interval of $(1.5,2.5)$ Hz, as the needle is vibrating at 2 $Hz$, $FFT$ and $iFFT$ refers to the fast Feurier transform and its inverse transform. The energy of the remained signal $\mathbf{S'}_{i,j}$ can be calculated using
\begin{equation}
    \mathbf{E'}_{i,j} = \sum\limits_{t=1}^{T}|\mathbf{S'}_{i,j,t}|^2,
\end{equation}
where $\mathbf{S'}_{i,j,t}$ is the value of the signal $\mathbf{S'}_{i,j}$ at the time $t$. 

\begin{figure}[ht!]
\centering
\includegraphics[width=0.5\textwidth]{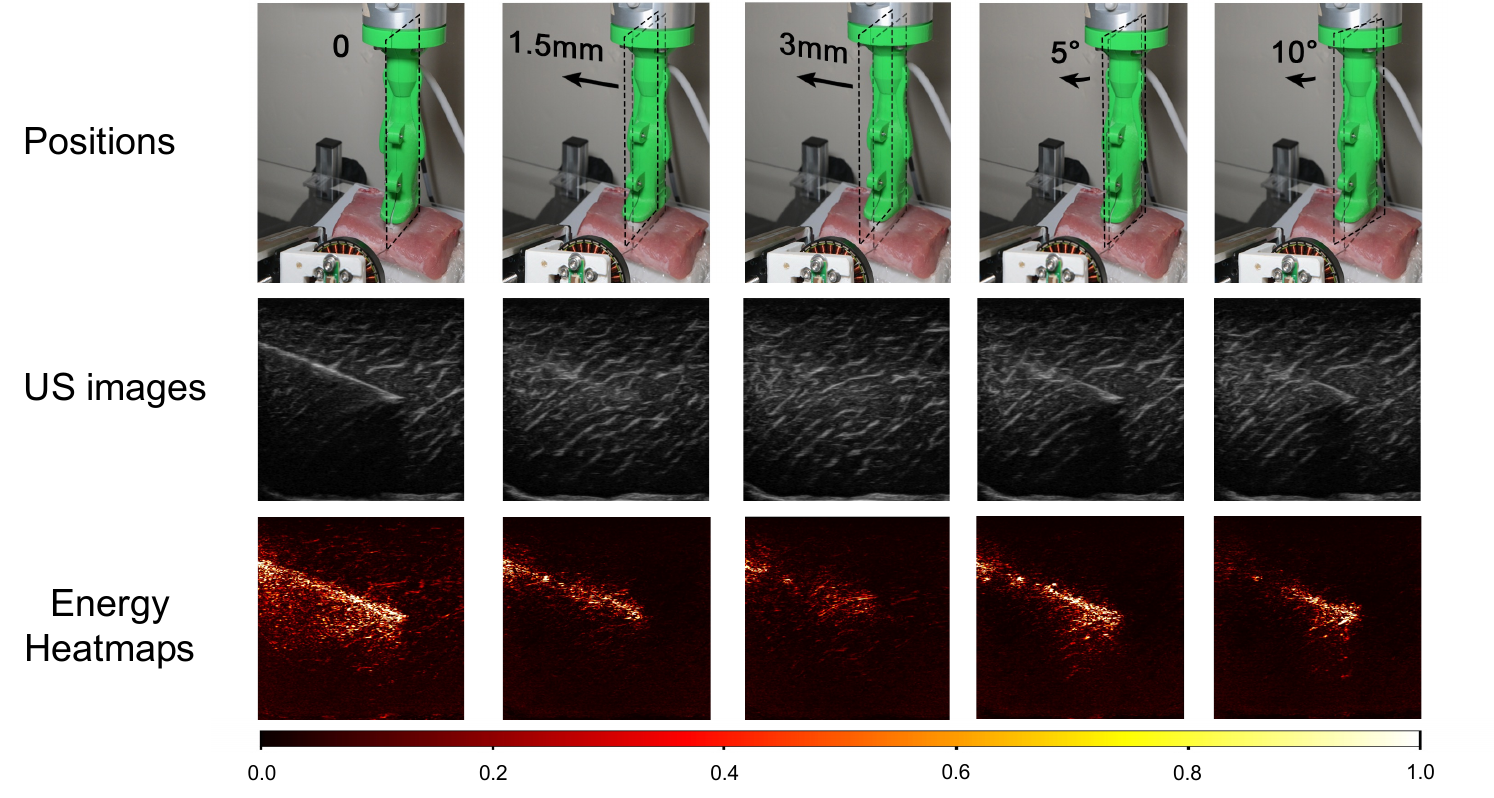}
\caption{Illustration of needle visibility in US slices and vibration-based energy heatmap with respect to varying translations and rotations. This representative result is obtained using a piece of porcine tissue. The upper row illustrates various positions of the ultrasound probe with translations or rotations from the needle plane; the middle row shows their corresponding US images; the last row shows vibration-based heatmaps of the vibration signal energies.
}
\label{fig_misalignment}
\end{figure}

The visualization of the results is shown in Fig.~\ref{fig_misalignment}. To enhance the visualizability, we removed a lot of values that are too small and normalized the images. The energy heatmaps visualize the vibration area and the vibration intensity in the imaging plane. When the imaging plane moves away from the needle, both the area and intensity decrease, which verifies our hypothesis. Especially when the translational offset is 3 mm, the needle is not shown in the US image, but the energy heatmap still highlights the region of the inserted needle. This shows the potential of restoring the needle even when the needle is not visible.

\subsection{Energy Metric} \label{attenuation}
\begin{figure}[ht!]
\centering
\includegraphics[width=0.48\textwidth]{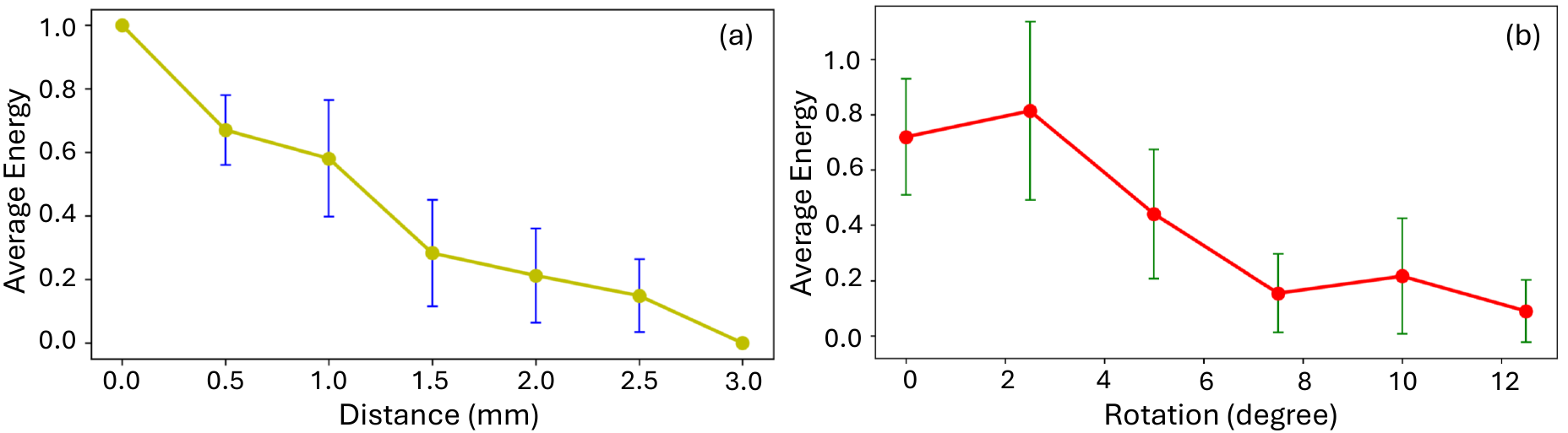}
\caption{Variation of the normalized average energy with translation and rotation relative to the needle plane. (a) is the variation of energy with distance, and (b) is the variation of energy with angle. The data was collected from 8 insertions on 3 different porcine tissue samples.}
\label{Fig_attenuation}
\end{figure}

In the vibration-based heatmap, regardless of the visibility of the needle in the ultrasound image, the needle and the range of the needle-driven vibration region can be clearly observed, as well as the decay of pixel intensity in these regions as the probe moves away from the vibration source. We introduce the average energy of each pixel in the US image to extract the relationship between the vibration-based ultrasound image and the probe offset:


\begin{equation}
    \mathbf{E}_{Avg} = \frac{\sum\mathbf{E'}_{i,j}}{H * W},
\end{equation}
Fig.~\ref{Fig_attenuation} depicts the normalized average energy per pixel in the ultrasound images as they change at different translation distances and rotation angles. The waypoints represent the average value of energy at each position, and the horizontal line represents the standard deviation. Although the energy decays more slowly and has larger fluctuations in the rotational case compared to the translational case, monotonic changes can be observed in both translation and rotation conditions. 

\section{ROBOTIC US PROBE REPOSITIONING}

\begin{figure}[ht!]
\centering
\includegraphics[width=0.5\textwidth]{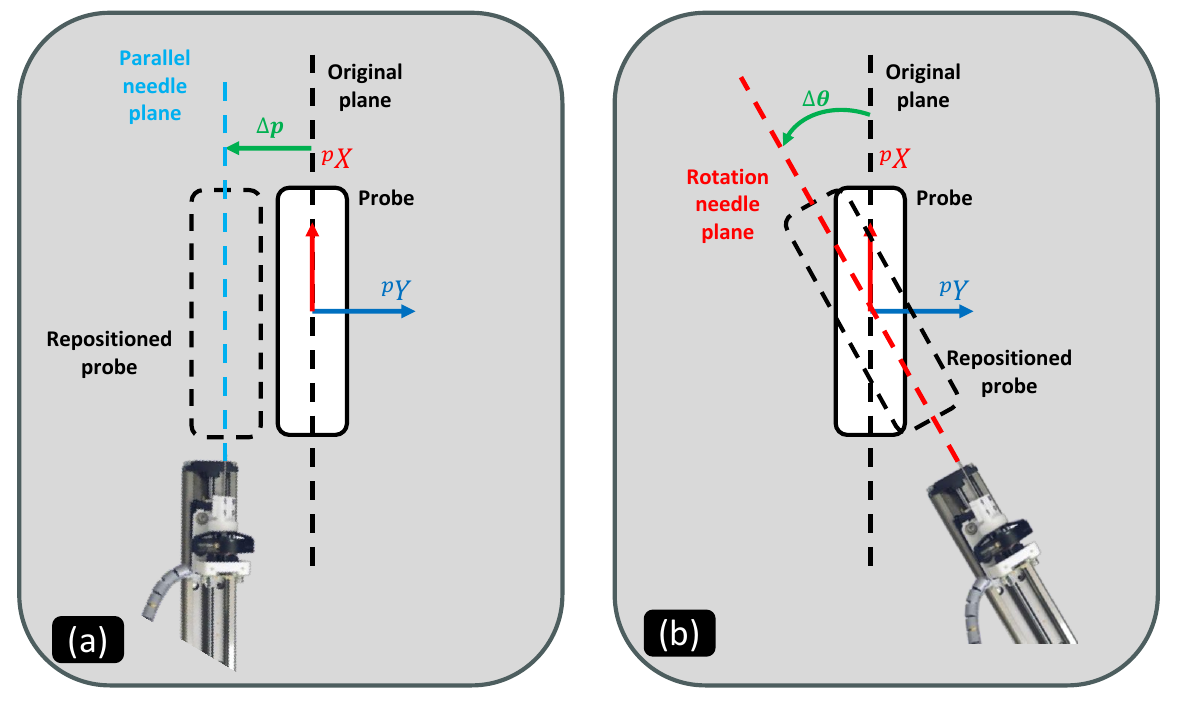}
\caption{Illustration of the probe coordinate and relative displacements. Top-down view along the $z$-axis of probe frame \{p\}. The solid and dashed rectangles represent the original and restored probe pose, respectively. $\Delta p$ and $\Delta \theta$ show the movement between the original position and the restored position.
}
\label{Fig_top_down}
\end{figure}

In this section, we present a control method for the autonomous repositioning of the US probe to restore the needle alignment in the US image. This control algorithm is driven by the metric of vibration-based energy. To demonstrate the feasibility, we study two representative cases of repositioning the US probe, i.e., the misalignments subject to translational offset and rotational offset, respectively, as shown in Fig.~\ref{Fig_top_down}. The local movements in translation and rotation are depicted in the probe frame $\{p\}$. $\Delta p$ is the displacement under parallel conditions and $\Delta \theta$ is the angle offset in the rotation condition. Our goal is to develop a strategy to mitigate the misalignment by maximizing the energy metric.

As mentioned in Section~\ref{attenuation}, the average energy after the bandpass filter increases as the misalignment between the ultrasound image plane and the needle insertion plane reduces. Inspired by the energy attenuation during the misalignment, using this displacement-related feature, by calculating the energy difference between two adjacent positions, it can provide direction and displacement guidance for the movement of the probe. Therefore, we propose a proportional control to achieve vibration-guided US probe repositioning for restoring needle alignment, as described in Algorithm \ref{alg:algorithm1}.  Specifically, in \textbf{Phase 1}, the probe first acquires an ultrasound image at the current position and calculates the average energy $E_{Avg,0}$. The robot then moves one predefined step $S$ and calculates the average energy $E_{Avg,1}$ in the same way. Thus, we can determine the direction, i.e., left or right for the translational case and clockwise or counter-clockwise for the rotational case. After determining direction, \textbf{Phase 2} can be conducted. The energy difference $E_{Avg, diff}$ corresponding to the two image sequences is used as the input of the proportional controller to control the continued movement of the probe. The parameter $K_p$ can be estimated from the slope of the fitting curves in Fig. \ref{Fig_attenuation} and can be fine-tuned in practice if necessary.  At the same time, we keep comparing the energy difference $E_{Avg,diff}$ and the energy threshold $T$ until the difference is smaller than the threshold to reposition the US probe for clear needle visibility. 
Moreover, we utilize a low-energy filter $E_{Avg, low}$ to avoid the navigation being stuck in the energy local maximum and improve movement efficiency. When the energy of the current position $E_{Avg,i+1}$ is smaller than the filter, the probe will keep moving. Here we use 20\% of the maximum energy as the filter by experience.

\begin{algorithm}
    \caption{Control algorithm for US probe repositioning}\label{alg:algorithm1}
    \begin{algorithmic}[1]
        \STATE Initialize
        \STATE $K_p \gets \text{Proportional gain}$
        \STATE $E_{Avg,diff} \gets \text{Initial energy difference}$
        \STATE $E_{Avg,low} \gets \text{Low-energy filter}$
        \STATE $H \gets \text{Energy calculation formula}$
        \STATE $S \gets \text{Step length used for determining direction}$
        \STATE $T \gets \text{Energy threshold}$
        \STATE $(F_{low}, F_{high}) \gets \text{Passband of bandpass filter}$
        \STATE $control\_output \gets 0$
        \STATE \textbf{Phase 1: Determine Direction}
            \STATE // Calculate energy of the starting position
            \STATE $E_{Avg,0} \gets H(I_{0})$
            \STATE // Probe moves one predefined step
            \STATE $control\_output \gets S$
            \STATE // Calculate energy after the step
            \STATE $E_{Avg,1} \gets H(I_{1})$
             \STATE // Calculate energy difference between two positions
             \STATE $E_{Avg,diff} \gets E_{Avg,0} - E_{Avg,1}$ 
             \STATE Determine direction based on the energy difference
        \STATE \textbf{Phase 2: Eliminate Misalignment}
        \WHILE{true}
            \STATE // Calculate energy of the previous position
            \STATE $E_{Avg,i} \gets H(I_{i})$
            \STATE // Calculate energy of the current position
            \STATE $E_{Avg,i+1} \gets H(I_{i+1})$
             \STATE // Calculate energy difference between two positions
             \STATE $E_{Avg,diff} \gets E_{Avg,i} - E_{Avg,i+1}$ 
            \IF{$E_{Avg,diff} < T$}
                \IF{$E_{Avg,i+1} > E_{Avg,low}$}
                    \STATE \textbf{Break}
                \ENDIF
            \ELSE
                \STATE // Calculate control output
                \STATE $control\_output \gets K_p \times E_{Avg,diff}$
            \ENDIF
        \ENDWHILE
    \end{algorithmic}
\end{algorithm}

\section{RESULTS}
\subsection{Experimental Setup}

In this study, all US data were acquired from an ACUSON Juniper US machine (Siemens Healthineers, Germany) using a linear probe 12L3 (Siemens Healthineers, Germany) with a footprint length of 51.3 mm. The US probe was mounted on the end-effector of a robotic arm (UR5e, Universal Robots, Denmark). We employed a self-developed Robot Operating System (ROS) interface~\cite{hennersperger2016towards} for all data acquisition and robotic system manipulation. The real-time US images were recorded by a frame grabber (Magewell, China) via OpenCV~\cite{bradski2000opencv}. To vibrate the inserted needle (18 G, length: 90 mm), we utilize an integrated vibration system with one brushless motor (Antigravity MN5008 KV170, T-motor, China) and controllers (MINI FSESC4.20, Flipsky, China). The motor is for the vibration around the shaft axis. We employed a single-board computer (Raspberry Pi 4B, UK) to control the motor system and communicate with the vibration system via the PC. Without further specification, the default needle's vibration only includes the vibration around the shaft. The motor periodically rotates 10 degrees clockwise and then 10 degrees counterclockwise at approximately 2 Hz. For the US image acquisition, the default settings on US machine are: image depth was 50 mm with a focus depth of 30 mm; tissue harmonic imaging (THI) was 6.7 MHz; dynamic range (DynR) was 80 dB.


\subsection{Robotic US Probe Restoration Performance}
\begin{figure}[ht!]
\centering
\includegraphics[width=0.5\textwidth]{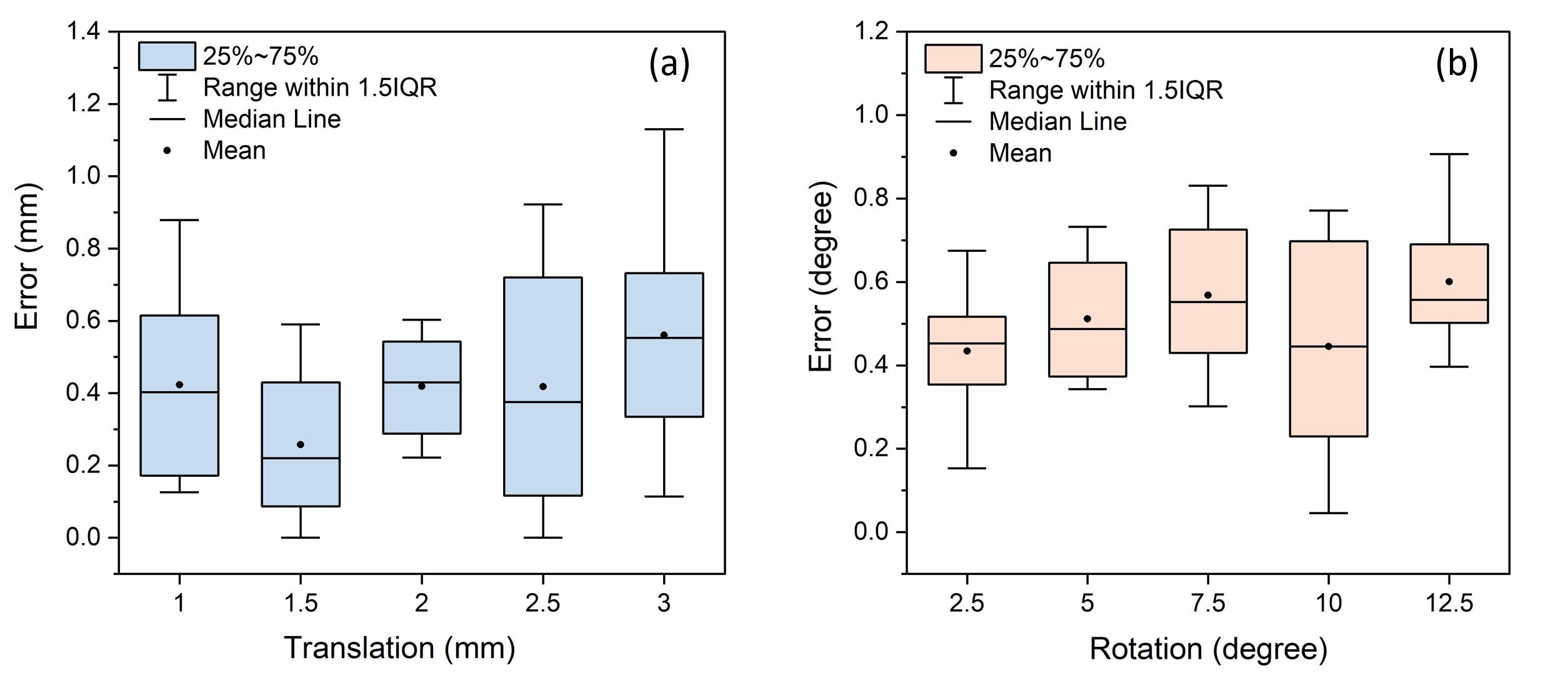}
\caption{Performance of the needle restoration on porcine tissue samples. (a) 
The probe was translated in the orthogonal direction with $\Delta p$. (b) The probe was rotated along its centerline in different angles $\Delta  \theta$.
}
\label{Fig_porcine}
\end{figure}

To evaluate the performance of the proposed method, we conducted experiments on two samples of porcine tissue for both translations and rotations. The influence of translation and rotation is disentangled by separately considering them. In each experiment, we identify the optimal imaging plane manually and translate the imaging plane to offset it by 1.0, 1.5, 2.0, 2.5, and 3.0 mm or rotate the probe 2.5, 5.0, 7.5, 10.0, and 12.5 degrees away from the best plane. After each offset, we run our algorithm to restore the imaging plane and measure the error between the best and the restored imaging plane. Fig.~\ref{Fig_porcine} shows the statistical results as a box plot. The average errors are less than 0.6 mm and 0.6 degrees in translation and rotation, respectively. We can also observe that the performance in translation is slightly better than in rotation. This occurrence can be attributed to the following fact: as shown in Figure~\ref{Fig_attenuation}, the feature of decreasing vibration energy becomes more pronounced when considering translation than when considering rotation as the offset increases. In the worst case, the distance of the restored imaging plane from the optimal imaging plane is 1.13 mm when the translation offset is set as 3 mm and is 0.91 degrees when rotating the probe 12.5 degrees from the base imaging plane. Even in the worst case, the needle can be clearly seen in the ultrasound images after the restoration, showing the correctness of the restoration. 



\begin{figure}[ht!]
\centering
\includegraphics[width=0.5\textwidth]{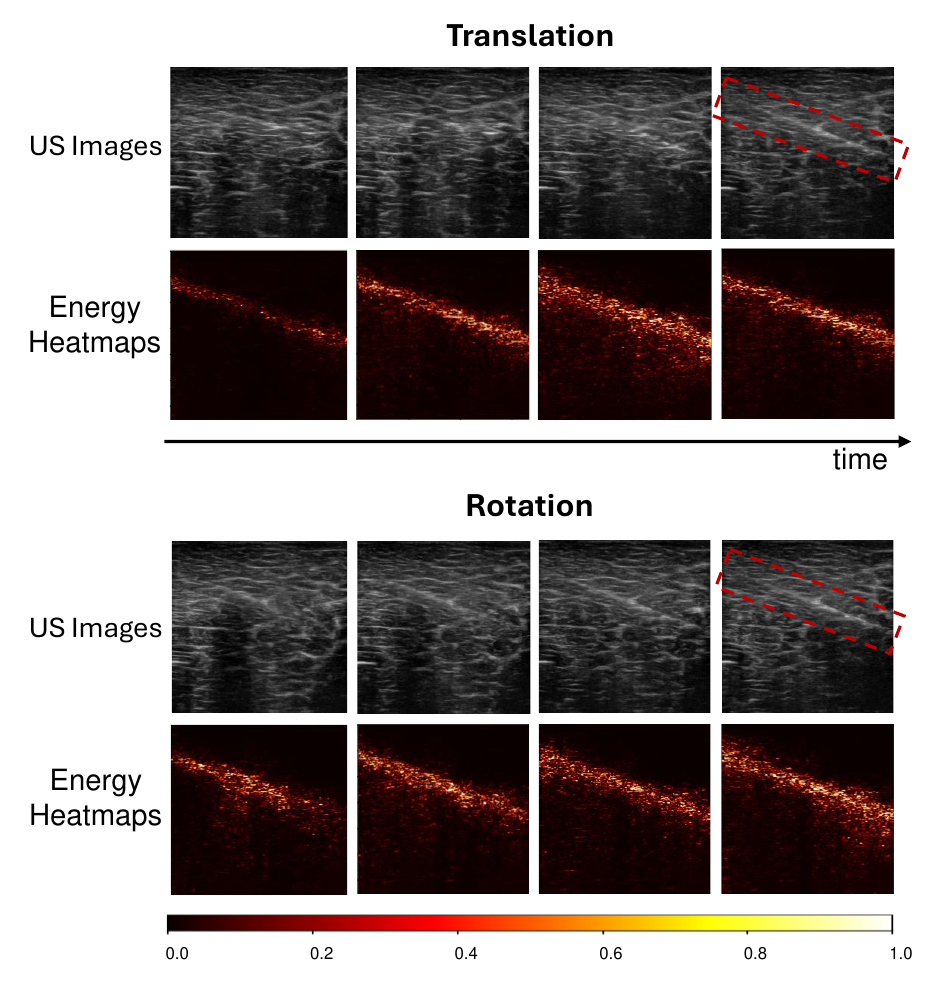}
\caption{Snapshots when running our algorithms to restore needle alignment. The upper image shows a time sequence of US images and the energy heatmaps during the restoration when considering translation (3 mm) as the cause of misalignment, while the misalignment in the bottom image is caused by rotation (12.5 degrees).
}
\label{Fig_process}
\end{figure}

Fig.~\ref{Fig_process} presents two sequences of snapshots captured during the needle alignment restoration process, covering both translational and rotational misalignments. Initially, the needle was difficult to localize in the ultrasound images due to misalignment. However, it remained detectable in the energy heatmaps generated from the vibration-based energy metric. Using this information, the ultrasound probe was repositioned accordingly to restore alignment. By the final snapshot, the needle was successfully tracked and became clearly visible in the ultrasound image, demonstrating the effectiveness of the proposed method in realigning the probe for accurate needle alignment.

\section{DISCUSSION AND CONCLUSION}
 
In this work, we proposed a new method to restore needle alignment using needle vibration, implemented within a dual-arm robotic needle insertion system equipped with a needle vibration device and intraoperative ultrasound guidance. Our approach enables stable and accurate ultrasound probe restoration for robust needle tracking. Through systematic experiments on ex-vivo porcine tissue samples, we analyzed vibration energy variations in ultrasound images across different translational offsets and rotational misalignments. Building on these findings, we proposed a vibration-based robotic probe repositioning method to restore needle alignment. To evaluate the system’s performance, we conducted restoration experiments under various translational and rotational offsets. The results demonstrated a translational error of 0.41 ± 0.27 mm and a rotational error of 0.51 ± 0.19 degrees. Within this error range, the needle remained clearly visible in the ultrasound image, validating the effectiveness of our approach. 

The main limitation of the proposed system is that it only supports pure translational and rotational probe movements, and its navigation speed is relatively slow. Although our study focuses on these constrained motion types, the results remain comparable to vision-based approaches such as \cite{jiang2024needle}, which reported similar errors in comparable settings. This highlights the potential of combining our vibration-based method with vision-based techniques to enhance needle alignment performance further. We also noticed that in Fig.~\ref{Fig_attenuation}, unlike the test results in translation, the best imaging plane of rotation did not obtain the heatmap with the maximum energy, which might come from measurement error. In addition, a higher energy is measured when the probe is rotated 10 degrees from the best imaging plane compared with 8 degrees. This may be because the imaging plane is still fully inside the needle at 8 degrees, while it is out of the needle at 10 degrees. More vibration energy can be detected from the surrounding tissues compared with imaging inside the needle. However, more investigation should be conducted to explain the energy metric. Furthermore, during large-scale translation and rotation movements, the change of vibration energy metric lacks clear features. These problems are all related to the generation and propagation of different vibration frequencies. The navigation speed is also limited by the frequency of the vibration. Therefore, the next step is to mine more information from the vibrating region of images and combine deep learning and dynamic vibration frequency to achieve efficient and stable needle tracking. Finally, our approach will be integrated into an RUSS designed for surgical applications, enabling real-time monitoring of the insertion process and the autonomous restoration of the visibility of the inserted instrument in case of misalignment.



\bibliography{references}

\end{document}